\documentclass[letterpaper, 10 pt, conference]{ieeeconf}

\IEEEoverridecommandlockouts
\usepackage{amsmath}
\usepackage{amssymb}
\usepackage{amsfonts}
\usepackage{bm}
\usepackage{graphicx}
\usepackage{cite}
\usepackage{booktabs}
\usepackage[table]{xcolor}
\usepackage{tabularx}
\usepackage{placeins}
\usepackage{textcomp}
\usepackage{url}

\definecolor{mygray}{gray}{.9}
\title{\LARGE \bf
 Configuration-Induced Passive Self-Rotation for\\
 Perception-Enhanced Autonomous Flight
}

\author{Xurui Liu, Miao Wang, Tianyu He, Linrui Yang, and Xiaobin Zhou$^{*}$%
\thanks{The authors are with the School of Robotics and Automation, Nanjing University, Suzhou 215163, China. (Emails: \mbox{xrl250444@gmail.com}, \mbox{wm\_ubuntu@nju.edu.cn}, \mbox{tonyhe\_2015321@outlook.com}, \mbox{linruiyang2004@163.com}, \mbox{xiaobin\_nju@nju.edu.cn})}%
\thanks{$^{*}$Corresponding author.}%
}

\begin{document}

\maketitle
\thispagestyle{empty}
\pagestyle{empty}
\suppressfloats[t]

\begin{abstract}
Autonomous flight in confined and cluttered environments is fundamentally limited by the restricted field of view (FoV) of onboard sensors. Passive self-rotation expands sensing coverage without additional sensors but introduces a tradeoff between swept-FoV refresh rate and flight performance. This letter presents a configuration-induced passively self-rotating tricopter for perception-enhanced autonomous flight. Firstly, the rear-arm configuration parameter is exploited to regulate the passive self-rotation operating point, providing an airframe-level mechanism for balancing swept-FoV refresh rate and flight performance. Secondly, a hierarchical autonomy framework integrating planning and control is developed to enable agile and robust autonomous flight under continuous passive self-rotation. For waypoint-based inspection, guide-point replanning is further used to improve task-level coverage. Extensive real-world experiments, including high-speed trajectory tracking, disturbance-rejection tests, and autonomous navigation in representative cluttered environments, demonstrate the effectiveness of the proposed approach for perception-enhanced autonomous flight.
\end{abstract}

\section{Introduction}

Small unmanned aerial vehicles (UAVs) are increasingly deployed in confined and cluttered environments, such as forests~\cite{lu2024gps}, underground garages and narrow passages~\cite{agha2022nebula,tranzatto2023cerberus}, and warehouses. Reliable navigation in these environments requires continuous obstacle perception and sufficiently complete local mapping. However, size, weight, and power constraints often limit small UAVs to compact sensors rigidly mounted on the airframe. Consequently, the observable region is restricted by the sensor field of view (FoV) and mounting direction, leaving lateral, rear, and near-ground regions under-observed. This limitation is particularly critical in cluttered environments, where incomplete observations can delay obstacle discovery, reduce planning efficiency, and increase navigation risk~\cite{chen2026pulsar2}.

\begin{figure}[t]
    \centering
    \includegraphics[width=0.85\columnwidth]{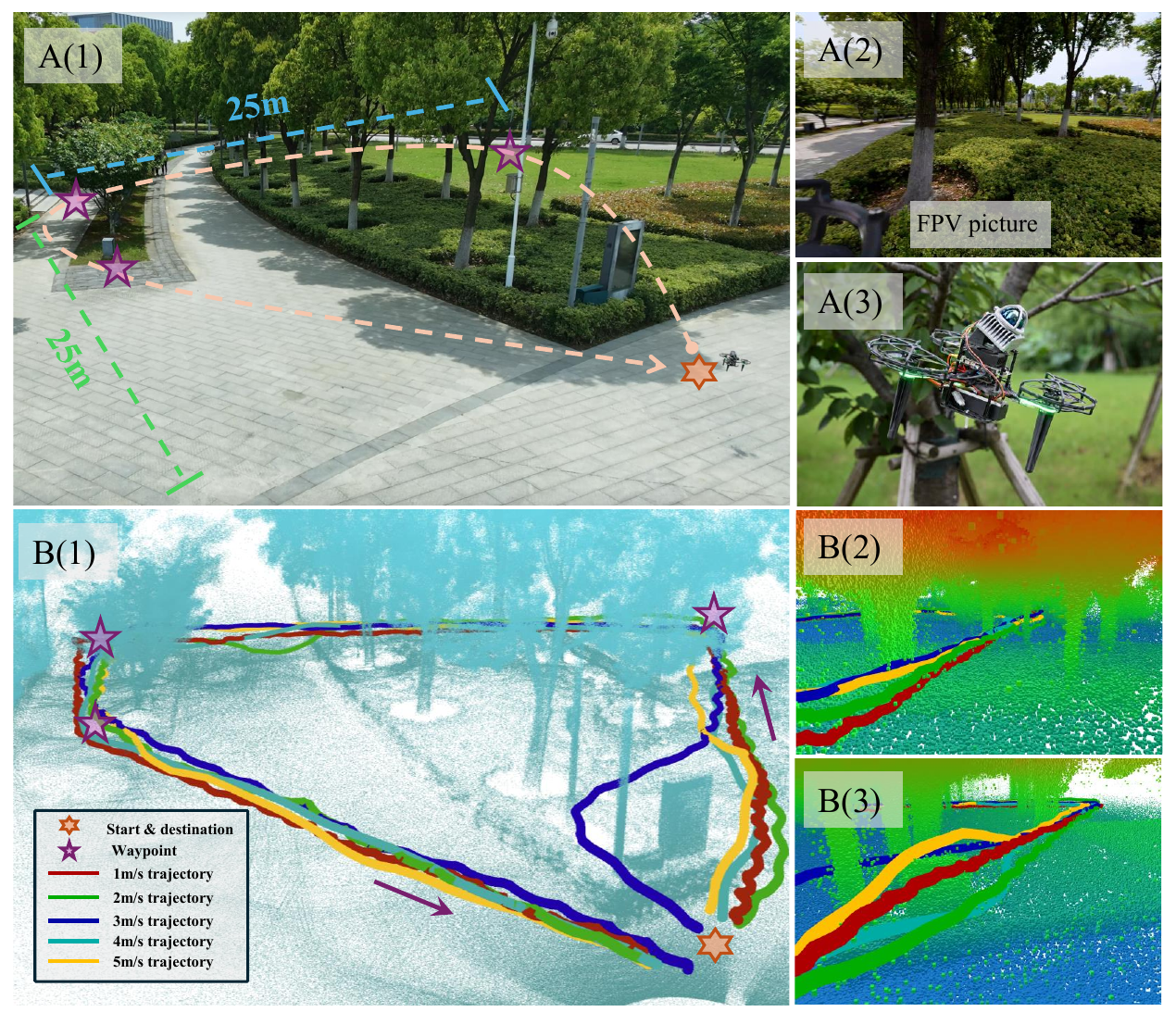}
    \caption{Autonomous navigation in an unknown forest environment. A(1) Bird's-eye view of the experimental environment, showing the flight area and waypoint layout. A(2) First-person view (FPV) captured during autonomous flight through a narrow passage. A(3) Chase-camera view of the tricopter during autonomous flight. B(1) Reconstructed point-cloud map overlaid with the start and destination, waypoints, and trajectories at $1$--$5\,\mathrm{m/s}$. B(2)-B(3) Local point-cloud view illustrating the planned trajectory through cluttered trees.}
    \label{fig:forest_navigation}
    \vspace{-1.5em}
\end{figure}

To mitigate under-observed regions caused by fixed and limited FoV, prior approaches enlarge the observable region using wide-FoV sensors~\cite{hao2021dual}, multiple sensors, active gimbals~\cite{chen2021active}, or vehicle self-rotation~\cite{cai2023pulsar,chen2026pulsar2,zhou2026spinner}. Additional sensors and active gimbals can improve coverage, but they introduce extra components, electromechanical degrees of freedom, or higher integration complexity, increasing the size, weight, and power burden on small UAVs. Vehicle self-rotation offers a different route by using body motion itself to expand the effective FoV, and recent self-rotating aerial vehicles have demonstrated enlarged sensing range and autonomous navigation in unknown environments~\cite{cai2023pulsar,chen2026pulsar2,zhou2026spinner}. However, these platforms mainly generate or regulate self-rotation through dedicated mechanisms, passive aerodynamic structures, or unconventional vehicle architectures, so the self-rotation rate is often treated as a platform-dependent property rather than an airframe-level design variable. Consequently, the relation among swept-FoV refresh, flight-control authority, and navigation performance remains insufficiently explored, especially because continuous self-rotation introduces attitude coupling, aerodynamic disturbances, and model mismatch during agile flight. These challenges motivate a unified framework that jointly considers platform design, flight control, and swept-FoV navigation for passively self-rotating aerial vehicles.

Motivated by these challenges, we develop a tricopter that induces passive self-rotation through its airframe configuration. The system couples airframe-level self-rotation-rate regulation, local replanning, and nonlinear model predictive control (NMPC) with incremental nonlinear dynamic inversion (INDI) compensation to support autonomous navigation under passive self-rotation (see Fig.~\ref{fig:forest_navigation}). For waypoint-based inspection, guide-point replanning uses swept-FoV map updates to prioritize less-observed regions. The main contributions of this work are summarized as follows:

\begin{enumerate}
    \item A configuration-induced passively self-rotating tricopter, shown in Fig.~\ref{fig:system_overview}, that enables airframe-level regulation of the passive self-rotation rate without requiring additional aerodynamic speed-regulation components.

    \item A perception-enhanced autonomous flight framework that combines local replanning and self-rotating trajectory tracking for robust navigation under passive self-rotation, with guide-point replanning for waypoint-based inspection.

    \item Extensive real-world experiments that validate the self-rotation mechanism, swept-FoV perception capability, high-precision trajectory tracking, and autonomous navigation performance in cluttered environments.
\end{enumerate}

\section{Related Work}

\subsection{Self-Rotating Aerial Vehicle Design and Control}

Self-rotating aerial vehicles have been explored as compact platforms that exploit continuous body rotation for flight and perception. The Monospinner demonstrated stable hovering with a continuously rotating body~\cite{mueller2016monospinner}, and recent single-actuator monocopter work achieved smooth trajectory tracking using INDI on Samara-inspired platforms~\cite{smooth2025monocopter}. Swashplateless micro aerial vehicles~\cite{powers2015swashplateless}, revolving-wing drones~\cite{chen2022bioinspired}, passive compliant variable-pitch wings~\cite{cai2023selfrotary}, and reconfigurable rotary-wing platforms~\cite{cai2025quadrotary} further showed that self-rotation, bio-inspired actuation, or reconfiguration can support efficient and agile flight. PULSAR and PULSAR II achieved omnidirectional sensing, autonomous navigation, agile tracking, and collision resistance on single-actuator platforms~\cite{cai2023pulsar,chen2026pulsar2}, while SPINNER validated vertical field-of-view expansion and autonomous flight with a passive tri-rotor configuration~\cite{zhou2026spinner}. However, in these platforms the self-rotation rate is mainly set by the vehicle architecture, flight mode, actuator dynamics, or auxiliary regulation elements rather than by a primary airframe design variable linked to swept-FoV refresh and flight-control authority. This work instead adjusts the passive self-rotation operating point through the tricopter airframe, balancing swept-FoV refresh with control authority without extra spin-regulation components.

\subsection{Perception-Enhanced Autonomous Navigation}

Autonomous flight in cluttered environments requires coupling perception, planning, and control under limited onboard sensing and computation~\cite{lu2024gps,liu2016highspeed,loianno2017estimation}. Perception-aware and information-driven planning methods improve navigation by considering sensing quality, informative viewpoints, or exploration frontiers during motion generation. Representative examples include perception-aware planning in unknown and feature-limited environments~\cite{yu2025perception}, perception-aware model predictive path integral control for unknown-environment navigation~\cite{zhai2026pamppi}, and exploration methods based on submaps, safe local replanning, unknown-region guidance, or lightweight LiDAR-based exploration~\cite{papatheodorou2025submap,oleynikova2018safe,liu2025flare,geng2025epic}.

The cited approaches obtain observations either from fixed or actively steered sensors, or from viewpoints explicitly selected during planning. In contrast, a passively self-rotating aerial vehicle naturally produces a time-varying swept FoV during flight. The resulting map updates support local replanning under passive self-rotation. In waypoint-based inspection, they are also used to select guide points toward less-observed regions.

\section{System Design and Configuration-Induced Self-Rotation}

\subsection{Hardware Platform and Software Architecture}

\begin{figure}[t]
    \centering
    \vspace*{6pt}
    \includegraphics[width=0.833\columnwidth]{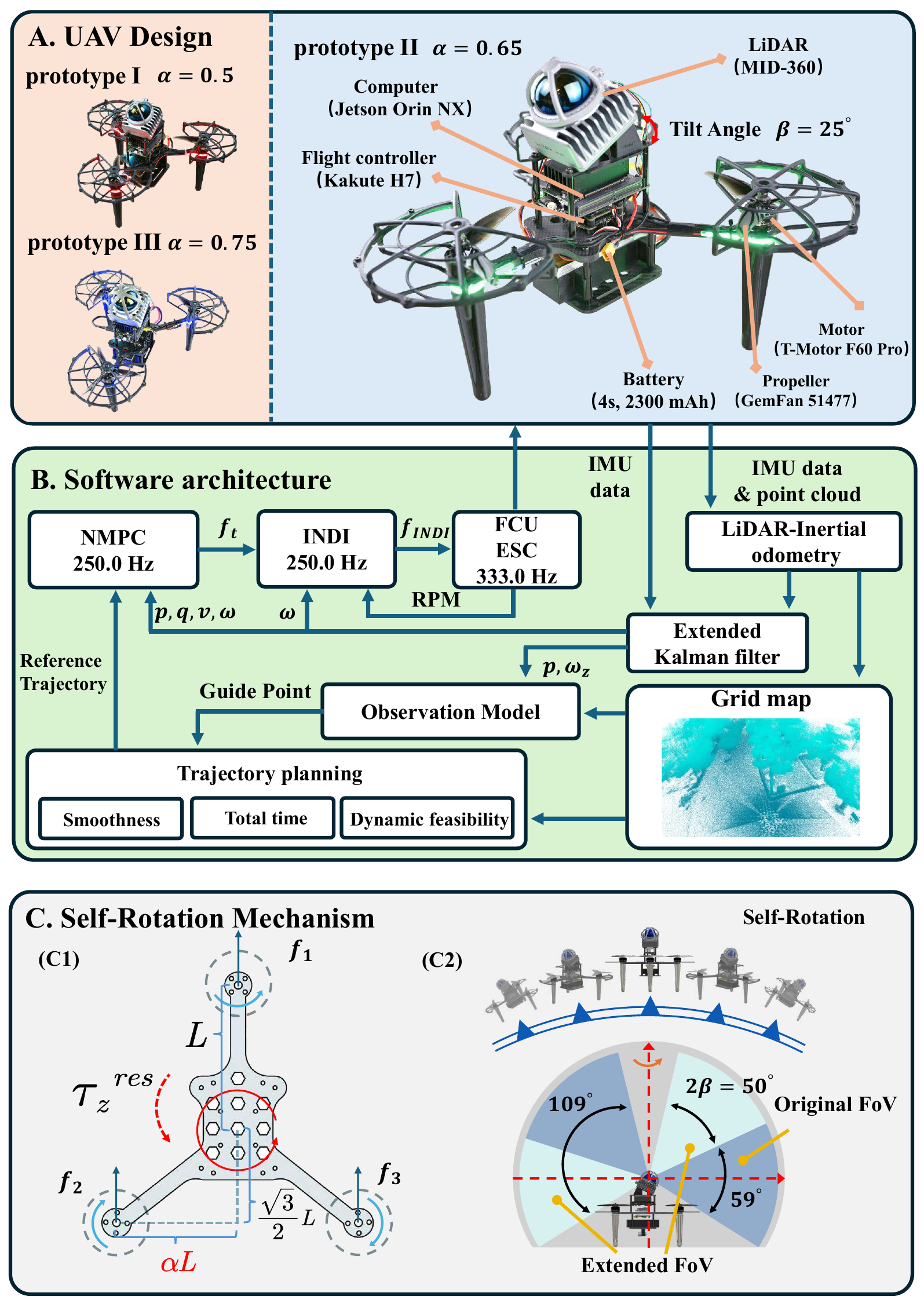}
    \caption{System overview of the configuration-induced passively self-rotating tricopter. (A) Three prototypes with different rear-arm configuration parameters $\alpha$, onboard hardware, and propulsion layout. (B) Onboard software architecture. (C) Self-rotation mechanism. (C1) Rear-arm geometry producing residual yaw counter-torque. (C2) Swept-FoV expansion from the native $59^\circ$ sensor FoV to an effective $109^\circ$ vertical envelope.}
    \label{fig:system_overview}
    \vspace{-1.5em}
\end{figure}

We treat the rear-arm configuration as an airframe-level design variable for setting the passive self-rotation operating point. As shown in Fig.~\ref{fig:system_overview}(A), we implement three tricopter prototypes that span different values of this design variable. Each prototype has one front arm and two symmetric rear arms, with rear-arm parameter $\alpha$ changing the rear-rotor pitch moment arm and regulating the steady passive self-rotation rate. To isolate the effect of $\alpha$, the prototypes share the same major hardware configuration. The onboard system integrates LiDAR sensing, onboard computing, flight control, and three-rotor propulsion modules. The LiDAR is mounted with a fixed tilt angle $\beta=25^\circ$, enabling vehicle self-rotation to transform the instantaneous FoV into a swept observation region. Fig.~\ref{fig:system_overview}(B) illustrates the onboard autonomy framework. LiDAR and inertial measurement unit (IMU) data are processed by FAST-LIVO2~\cite{zheng2025fastlivo2} and an extended Kalman filter for state estimation and local mapping. The resulting map supports local replanning and, for inspection, guide-point generation toward less-observed regions. The planner produces dynamically feasible reference trajectories, which are tracked by NMPC with INDI-based compensation. The compensated commands are executed by the flight-control unit (FCU) and electronic speed controllers (ESCs), with rotor-speed feedback returned to the INDI layer.

\subsection{Configuration-Induced Passive Self-Rotation}
\label{sec:configuration_self_rotation}

Fig.~\ref{fig:system_overview}(C1) illustrates how the rear-arm configuration induces passive self-rotation. Varying $\alpha$ changes the rear-rotor pitch moment arm, so balancing roll and pitch moments redistributes the rotor thrusts and alters the residual yaw counter-torque that drives passive self-rotation.

The three rotor thrusts are denoted by $\bm{f}=[f_1,f_2,f_3]^{\mathrm T}$, where each $f_i=k_t\varpi_i^2$ with $k_t$ the thrust coefficient and $\varpi_i$ the angular speed of rotor $i$. The rotor thrust-to-wrench mapping is
\begin{equation}
\begin{bmatrix}
F\\
\tau_x\\
\tau_y\\
\tau_z
\end{bmatrix}
=
\underbrace{
\begin{bmatrix}
1 & 1 & 1\\
0 & \frac{\sqrt{3}}{2}L & -\frac{\sqrt{3}}{2}L\\
-L & \alpha L & \alpha L\\
-\kappa_\tau & \kappa_\tau & \kappa_\tau
\end{bmatrix}}_{\bm{M}(\alpha)}
\bm{f} ,
\end{equation}
where $\bm{M}(\alpha)$ is the thrust allocation matrix, $L$ is the front-rotor pitch-axis moment arm, $\alpha L$ is the rear-rotor pitch-axis moment arm, $\kappa_\tau$ is the rotor counter-torque coefficient, and $m$ and $g$ are the platform mass and gravitational acceleration, respectively. Under hovering equilibrium and neglecting aerodynamic disturbances, the rotor thrusts satisfy
\begin{equation}
f_1=\frac{\alpha mg}{1+\alpha},
\qquad
f_2=f_3=\frac{mg}{2(1+\alpha)}.
\end{equation}
Substituting these thrusts into the yaw-moment row gives the residual yaw counter-torque
\begin{equation}
\tau_z^{\mathrm{res}}(\alpha)
=
\kappa_\tau(-f_1+f_2+f_3)
=
\kappa_\tau mg\frac{1-\alpha}{1+\alpha}.
\label{eq:residual_yaw_torque}
\end{equation}
Here, $\tau_z^{\mathrm{res}}(\alpha)$ denotes the net yaw moment generated by the equilibrium thrust distribution. For $\alpha<1$, this residual yaw counter-torque decreases as $\alpha$ approaches one, so a smaller $\alpha$ provides a larger driving moment for passive self-rotation. This relation links the rear-arm configuration to the passive self-rotation operating point. Since the steady self-rotation rate is set by the balance between this driving moment and aerodynamic rotational damping, it is determined experimentally through steady-hover tests.

The perception effect is illustrated in Fig.~\ref{fig:system_overview}(C2). Continuous passive self-rotation converts the instantaneous LiDAR FoV into a swept-FoV envelope. The native vertical FoV of the Livox Mid-360 is $\mathrm{FoV}_{\mathrm{native}}=59^\circ$. Over one self-rotation period, the accumulated vertical angular envelope can be approximated as
\begin{equation}
\mathrm{FoV}_{\mathrm{eff}}
\approx
\mathrm{FoV}_{\mathrm{native}}+2\beta
=109^\circ.
\end{equation}

\section{Self-Rotating Flight Dynamics and Model Identification}

\subsection{Self-Rotating Flight Dynamics Model}

The model is expressed in a world frame $W$ and a body-fixed frame $B$. The vehicle state consists of the world-frame position $\bm{p}$, world-frame velocity $\bm{v}$, attitude quaternion $\bm{q}$, and body-frame angular velocity $\bm{\omega}$. The rotation matrix from $B$ to $W$ is denoted by $\bm{R}$, and $\bm{e}_3=[0,0,1]^{\mathrm T}$ denotes the vertical unit vector. The product $\bm{R}\bm{e}_3$ gives the body thrust-axis direction in $W$. The self-rotating flight model is
\begin{equation}
m\ddot{\bm{p}}
=
F\bm{R}\bm{e}_3
-
\bm{R}\bm{D}\bm{R}^{\mathrm T}\bm{v}
-
mg\bm{e}_3 ,
\end{equation}
\begin{equation}
\dot{\bm{q}}
=
\frac{1}{2}
\bm{q}
\otimes
\begin{bmatrix}
0\\
\bm{\omega}
\end{bmatrix},
\end{equation}
\begin{equation}
\dot{\bm{\omega}}
=
\bm{J}^{-1}
\left(
\bm{\tau}
+
\bm{\tau}_{\mathrm{ip}}
-
\bm{\omega}\times\bm{J}\bm{\omega}
-
\bm{\tau}_{\mathrm{aero}}
\right).
\end{equation}
Here, $\bm{D}$ is the translational drag matrix, $\bm{\tau}$ is the rotor-generated body moment, $\bm{\tau}_{\mathrm{ip}}$ denotes the moment induced by rotor inertia, and $\bm{\tau}_{\mathrm{aero}}$ denotes the lumped rotational residual during self-rotating flight. The symbols $\otimes$ and $\times$ denote quaternion multiplication and the vector cross product, respectively. Motivated by the lumped rotor-drag formulation for high-speed multirotor flight~\cite{faessler2018differential}, this residual is approximated by a linear model,
\begin{equation}
\bm{\tau}_{\mathrm{aero}}
\approx
\bm{c}
+
A_v\bm{R}^{\mathrm T}\bm{v}
+
B_{\omega}\bm{\omega},
\label{eq:rotational_residual_model}
\end{equation}
where $\bm{c}$ is a residual bias vector, and $A_v$ and $B_{\omega}$ are $3\times3$ residual-moment coefficient matrices associated with body-frame velocity and angular-rate effects. The three rows correspond to roll, pitch, and yaw residual moments.

\subsection{Model Parameter Identification}

We measure $m$ directly, obtain $\bm{J}$ from the assembled CAD model, and calibrate $k_t$ and $\kappa_\tau$ through static bench measurements.

The translational drag coefficient is identified from flight data. Let $\bm{v}_B=\bm{R}^{\mathrm T}\bm{v}$ be the body-frame velocity. From the translational dynamics, the acceleration residual associated with translational drag is written as
\begin{equation}
\bm{\eta}_D
=
\bm{R}^{\mathrm T}
\left(
\frac{F}{m}\bm{R}\bm{e}_3
-
g\bm{e}_3
-
\ddot{\bm{p}}
\right)
=
\frac{\bm{D}}{m}\bm{v}_B .
\end{equation}
The entries of $\bm{D}/m$ are obtained by linear least-squares regression between $\bm{\eta}_D$ and $\bm{v}_B$.

We identify the rotational residual from lemniscate flight logs collected at multiple reference speeds. The residual is constructed as
\begin{equation}
\bm{\eta}_{\omega}
=
\bm{\tau}
+
\bm{\tau}_{\mathrm{ip}}
-
\bm{J}\dot{\bm{\omega}}
-
\bm{\omega}\times\bm{J}\bm{\omega}.
\end{equation}
$\bm{c}$, $A_v$, and $B_{\omega}$ are estimated by fitting \eqref{eq:rotational_residual_model} to the rotational residual using flight segments with sufficient translational and rotational excitation. The fitted coefficients help the nominal model capture prototype-specific rotational residuals.

\section{Controller Design and Guide-Point Replanning}

\subsection{Tracking Control with Disturbance Compensation}

At each control cycle, the NMPC optimizes a rotor-thrust command sequence over horizon $N$ with step size $\Delta t$. The prediction state augments the rigid-body state with the actual rotor thrusts,
\begin{equation}
\bm{x}
=
[\bm{p}^{\mathrm T},\bm{q}^{\mathrm T},\bm{v}^{\mathrm T},\bm{\omega}^{\mathrm T},\bm{f}^{\mathrm T}]^{\mathrm T},
\qquad
\dot{\bm{f}}
=
\frac{1}{\tau_m}
(\bm{u}-\bm{f}),
\label{eq:nmpc_thrust_dynamics}
\end{equation}
where $\bm{u}$ is the thrust command and $\tau_m$ is the motor time constant. The transition map $\Phi$ used in the NMPC constraints is obtained by discretizing the rigid-body dynamics and \eqref{eq:nmpc_thrust_dynamics}. Under passive self-rotation, yaw evolves with the vehicle's natural spinning motion. The tracking-error vector is therefore defined so that the NMPC cost penalizes thrust-axis misalignment rather than full attitude error,
\begin{equation}
\bm{y}_i
=
\begin{bmatrix}
\bm{p}_i-\bm{p}_i^{\rm ref}\\
\left(\bm{R}_i\bm{e}_3\times\bm{b}_{3,i}^{\rm ref}\right)_{xy}\\
\bm{v}_i-\bm{v}_i^{\rm ref}\\
\bm{\omega}_{xy,i}-\bm{\omega}_{xy,i}^{\rm ref}
\end{bmatrix},
\end{equation}
where $\bm{\omega}_{xy}=[\omega_x,\omega_y]^{\mathrm T}$ and $\bm{b}_{3,i}^{\rm ref}$ is the reference thrust axis. Let $\Delta\bm{f}_i=\bm{f}_i-\bm{f}_i^{\rm ref}$, $\Delta\bm{u}_i=\bm{u}_i-\bm{u}_i^{\rm ref}$, and $\bm{z}_i=[\bm{y}_i^{\mathrm T},\Delta\bm{f}_i^{\mathrm T}]^{\mathrm T}$. The NMPC problem is
\begin{align}
\bm{u}_{k:k+N-1}^{\ast}
=
&\arg\min_{\bm{u}_k,\ldots,\bm{u}_{k+N-1}}
\bm{z}_{k+N}^{\mathrm T}\bm{Q}_N\bm{z}_{k+N}
\nonumber\\
&+
\sum_{i=k}^{k+N-1}
\left(
\bm{z}_i^{\mathrm T}\bm{Q}\bm{z}_i
+
\Delta\bm{u}_i^{\mathrm T}
\bm{R}_u
\Delta\bm{u}_i
\right)
\end{align}
subject to
\begin{equation}
\bm{x}_{i+1}=\Phi(\bm{x}_i,\bm{u}_i),
\qquad
\bm{u}_{\min}\leq\bm{u}_i\leq\bm{u}_{\max},
\end{equation}
for $i=k,\ldots,k+N-1$. The state-error weight follows the ordering of $\bm{z}_i$ as $\bm{Q}=\mathrm{diag}(\bm{Q}_p,\bm{Q}_{\rm tilt},\bm{Q}_v,\bm{Q}_{\omega},\bm{Q}_f)$. The matrix $\bm{R}_u$ weights the thrust-command deviation, and $\bm{Q}_N$ weights the terminal cost.

We use INDI~\cite{tal2021accurate} to compensate residual aerodynamic and motor-response errors. The subscript $xy$ denotes the roll-pitch components, with yaw left to the passive self-rotation dynamics. The feedback terms $\bm{\tau}_{xy}^{\rm f}$ and $\dot{\bm{\omega}}_{xy}^{\rm f}$ are obtained from thrust estimates derived from rotor speeds and from IMU angular rate measurements after low pass filtering, respectively. The desired lateral moment is
\begin{equation}
\bm{\tau}_{xy}^{\rm d}
=
\bm{\tau}_{xy}^{\rm f}
+
\bm{J}_{xy}
\left(
\dot{\bm{\omega}}_{xy}^{\rm des}
-
\dot{\bm{\omega}}_{xy}^{\rm f}
\right),
\end{equation}
where $\dot{\bm{\omega}}_{xy}^{\rm des}$ is the desired roll-pitch angular acceleration and $\bm{J}_{xy}$ is the roll-pitch inertia submatrix. The INDI-compensated rotor thrust command is

\begin{equation}
\bm{u}_{\mathrm{INDI}}
=
\bm{M}_{Fxy}^{-1}(\alpha)
\begin{bmatrix}
F\\
\bm{\tau}_{xy}^{\rm d}
\end{bmatrix}.
\end{equation}
Here, $\bm{M}_{Fxy}(\alpha)$ contains the total-thrust, roll-moment, and pitch-moment rows of $\bm{M}(\alpha)$.

\subsection{Guide-Point Selection and Replanning}

Passive self-rotation continuously produces swept-FoV map updates. At replanning cycle $n$, these updates are used to form the observed free space and its frontier boundary. Let $\theta_{S,k}(\mathbf x)$ be the vertical angle of voxel $\mathbf x$ in the tilted LiDAR frame at observation update $k$, computed using the vehicle yaw and LiDAR tilt $\beta$, and let $\operatorname{LOS}_k(\mathbf x)=1$ indicate an unobstructed line of sight (LOS) verified by ray casting. The observed set, safe free candidate set, and frontier set are
\begin{align}
\mathcal O_n
&=
\bigcup_{k\in\mathcal K_n}
\{\mathbf x\mid
r_{\min}\le\|\mathbf x-\mathbf p_k\|\le R_\ell,
\nonumber\\
&\qquad
\theta_{\ell}\le\theta_{S,k}(\mathbf x)\le\theta_u,
\operatorname{LOS}_k(\mathbf x)=1
\},
\nonumber\\
\mathcal X_n
&=
\{\mathbf x\in\mathcal O_n\mid \mathcal I_n(\mathbf x)=0\},
\nonumber\\
\mathcal B_n
&=
\{\mathbf b\in\mathcal X_n\mid
\exists \mathbf b'\in\mathcal N_6(\mathbf b),
\mathbf b'\notin\mathcal O_n
 \}.
\end{align}
Here, $\mathcal K_n$ is the update set before cycle $n$, $\mathbf p_k$ is the vehicle position at update $k$, $r_{\min}$ and $R_\ell$ are LiDAR range limits, and $\theta_\ell$ and $\theta_u$ are vertical FoV bounds. The condition $\mathcal I_n(\mathbf x)=0$ marks a collision-free voxel, and $\mathcal N_6(\mathbf b)$ gives the six-connected neighbors used to extract $\mathcal B_n$.

The guide point is searched in a sector window $\mathcal W_n$ along the current waypoint direction, with radius $R_s=\kappa_f H$, angular width $\psi_s$, and height threshold $h_s$. Let $\mathcal X_{\mathcal W}=\mathcal X_n\cap\mathcal W_n$ and $\mathcal B_{\mathcal W}=\mathcal B_n\cap\mathcal W_n$. For $\mathbf x\in\mathcal X_{\mathcal W}$, the score is
\begin{align}
J_g(\mathbf x)
=&\,
I_g(\mathbf x)
+
\lambda_p \mathbf d_n^{\mathrm T}(\mathbf x-\mathbf p_n)
\nonumber\\
&+
\lambda_\perp
\left\|
\left(\mathbf I-\mathbf d_n\mathbf d_n^{\mathrm T}\right)
(\mathbf x-\mathbf p_n)
\right\|,
\nonumber\\
I_g(\mathbf x)=
&\sum_{\substack{\mathbf b_j\in\mathcal B_{\mathcal W}\\ \rho_j\le R_\ell}}
\exp
\left(
-\frac{(\rho_j-R_K)^2}{2\sigma^2}
\right)
\left(1-\cos^2\theta_j\right).
\end{align}
Here, $\mathbf d_n$ is the current waypoint direction, while $\rho_j$ and $\theta_j$ are the distance and angular relation between candidate $\mathbf x$ and frontier voxel $\mathbf b_j$. Let $\mathbf x_{g,n}$ be the highest-scoring candidate at cycle $n$. The guide-point latch and local target are
\begin{align}
\eta_n
&=
\begin{cases}
1, &
\mathbf x_{g,n-1}^{\rm act}\ \text{is inactive or }
\|\mathbf p_n-\mathbf x_{g,n-1}^{\rm act}\|\le d_g,\\
0, & \text{otherwise},
\end{cases}
\nonumber\\
\mathbf x_{g,n}^{\rm act}
&=
\begin{cases}
\mathbf x_{g,n}, &
I_g(\mathbf x_{g,n})>0,\ \eta_n=1,\\
\mathbf x_{g,n-1}^{\rm act}, & \text{otherwise},
\end{cases}
\nonumber\\
\mathbf g_n
&=
\begin{cases}
\mathbf x_{g,n}^{\rm act}, &
\mathbf x_{g,n}^{\rm act}\ \text{is active and }
\|\mathbf w_n-\mathbf p_n\|>d_{\rm term},\\
\mathbf w_n, & \text{otherwise}.
\end{cases}
\end{align}
Given $\mathbf g_n$, a grid-based A* search finds a collision-free initial path from $\mathbf p_n$ to $\mathbf g_n$ on the local occupancy map, which minimum-control-effort (MINCO) optimization~\cite{wang2022geometrically} refines into a dynamically feasible trajectory.

\FloatBarrier
\section{Experiments}
\label{sec:experiments}

We evaluate the proposed system through a series of real-world experiments. These experiments examine whether passive self-rotation expands the effective vertical sensing coverage achieved with the tilted LiDAR (Section~VI-A), how the rear-arm configuration affects the passive self-rotation rate and flight agility (Section~VI-B), whether the identified model improves prediction accuracy and supports robust control under external disturbances (Sections~VI-C and VI-D), and evaluate sparse-waypoint navigation and guide-point replanning for relative coverage in waypoint-based inspection (Section~VI-E). Table~\ref{tab:experiment_parameters} lists the main experimental parameters for prototype II, selected in Section~VI-B. The nonlinear MPC problem is solved using the ACADO solver in conjunction with qpOASES, and tracking performance is evaluated by the position root-mean-square error (RMSE).
\vspace{-1.0em}
\begin{table}[!h]
\caption{Experimental Parameters for Prototype II}
\label{tab:experiment_parameters}
\centering
\begingroup
\scriptsize
\renewcommand{\arraystretch}{1.3}
\setlength{\tabcolsep}{3pt}
\resizebox{\columnwidth}{!}{%
\begin{tabular}{ll}
\toprule
Parameters & Values\\
\midrule
\rowcolor{mygray} $m$ [$\mathrm{kg}$], $L$ [$\mathrm{m}$] & $1.1245,\ 0.125$\\
$\bm{J}$ [$10^{-3}\,\mathrm{kg\,m^2}$], $\kappa_\tau$ [$\mathrm{m}$] & $\operatorname{diag}(5.90,5.43,4.65),\ 0.0125$\\
\rowcolor{mygray} $N$, $\Delta t$ [$\mathrm{s}$], $\tau_m$ [$\mathrm{s}$] & $20,\ 0.05,\ 0.02$\\
$\bm{Q}_p$, $\bm{Q}_v$ & $\operatorname{diag}(1000,1000,1000),\ \operatorname{diag}(1,1,1)$\\
\rowcolor{mygray} $\bm{Q}_{\rm tilt}$, $\bm{Q}_{\omega}$ & $\operatorname{diag}(10,10),\ \operatorname{diag}(1,1)$\\
$\bm{Q}_f$, $\bm{R}_u$ & $\operatorname{diag}(1,1,1),\ \operatorname{diag}(2,2,2)$\\
\rowcolor{mygray} $u_{\min}$, $u_{\max}$ [$\mathrm{N}$] & $0.001,\ 7.0$\\
$r_{\min}$ [$\mathrm{m}$], $R_\ell$ [$\mathrm{m}$] & $0.1,\ 9$\\
\rowcolor{mygray} $\theta_{\ell}$, $\theta_u$ & $-7^\circ,\ 52^\circ$\\
$H$ [$\mathrm{m}$], $\kappa_f$, $\psi_s$, $h_s$ [$\mathrm{m}$] & $10,\ 1,\ 150^\circ,\ 0.1$\\
\rowcolor{mygray} $R_K$ [$\mathrm{m}$], $\sigma$ [$\mathrm{m}$], $\lambda_p$, $\lambda_\perp$ & $3,\ 0.9,\ 0.5,\ 3$\\
$d_g$ [$\mathrm{m}$], $d_{\rm term}$ [$\mathrm{m}$] & $5,\ 7$\\
\bottomrule
\end{tabular}
}
\endgroup
\vspace{-1.5em}
\end{table}

\subsection{Swept-FoV Expansion}

\begin{figure}[h]
    \centering
    \includegraphics[width=0.85\columnwidth]{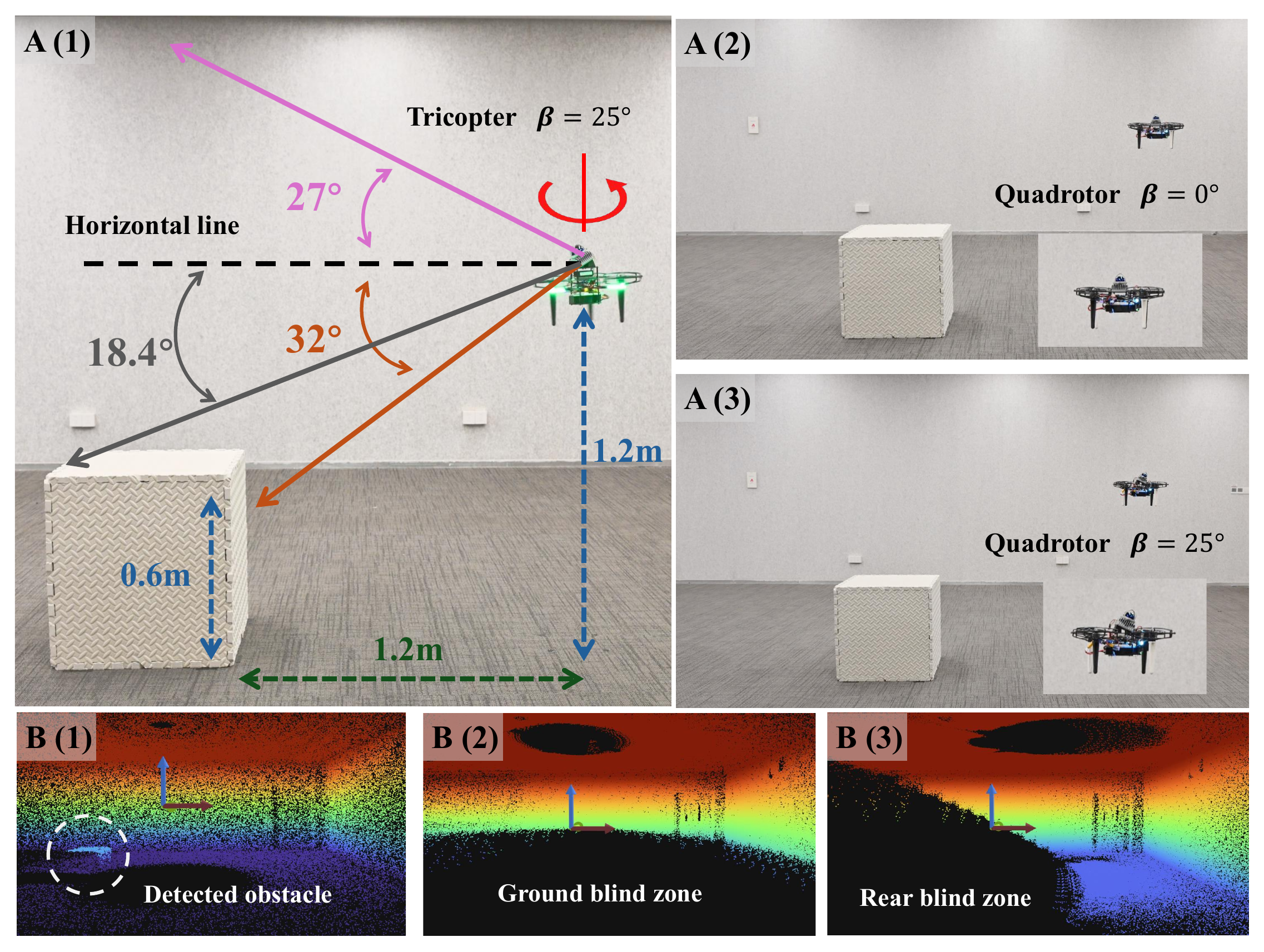}
    \caption{Swept-FoV expansion experiment. A(1) shows the tilted LiDAR geometry and rear obstacle. A(2)--A(3) show the horizontal and tilted fixed LiDAR references. B(1)--B(3) compare the corresponding point cloud observations.}
    \label{fig:perception_exp}
    \vspace{-0.5em}
\end{figure}

We compare the proposed self-rotating tricopter with two fixed-mount quadrotor references, as shown in Fig.~\ref{fig:perception_exp}. All vehicles hover at $1.2\,\mathrm{m}$, with a $0.6\,\mathrm{m}$ cubic obstacle placed $1.2\,\mathrm{m}$ away at a rear bearing. The evaluated obstacle edge in Fig.~\ref{fig:perception_exp}(A1) requires a depression angle of at least $18.4^\circ$. The horizontal reference provides only $7^\circ$ downward coverage, while the tilted reference reaches $32^\circ$ but only within a fixed forward sector. Both references therefore miss the rear obstacle. In contrast, passive self-rotation sweeps the vertical FoV of the tilted LiDAR around the vehicle, allowing the obstacle to be captured in the accumulated point cloud. This confirms vertical FoV expansion under passive self-rotation.
\vspace{-1.0em}
\begin{table*}[t]
\vspace*{9pt}
\caption{Prototype Selection Metrics, Lemniscate Tracking, and Gate Passing Results}
\label{tab:self_rotation_tracking}
\centering
\scriptsize
\renewcommand{\arraystretch}{1.05}
\setlength{\tabcolsep}{3pt}
\begin{tabular*}{\textwidth}{@{\extracolsep{\fill}} l c c c c c c c c c c}
\toprule
 & & & $\bar{\Omega}_z\pm\sigma$ & $T_s$ & $\Delta f_{\mathrm{RMS}}^{5\,\mathrm{m/s}}$ $\downarrow$ & $U_{95}^{5\,\mathrm{m/s}}$ $\downarrow$ & \multicolumn{4}{c}{$e_{\mathrm{RMSE}}$ $\downarrow$ ($\mathrm{m}$)} \\
\cmidrule(lr){8-11}
Task & $\alpha$ & $\tau_{\mathrm n}$ & ($\mathrm{rad/s}$) & ($\mathrm{s}$) & ($\mathrm{N}$) &  & $2\,\mathrm{m/s}$ & $3\,\mathrm{m/s}$ & $4\,\mathrm{m/s}$ & $5\,\mathrm{m/s}$ \\
\midrule
Eight (prototype I) & 0.50 & 0.333 & $11.03\pm0.28$ & 0.57 & 0.943 & \textbf{0.814} & 0.084 & 0.156 & 0.211 & 0.241 \\
Eight (prototype II) & 0.65 & 0.212 & $8.20\pm0.49$ & 0.77 & \textbf{0.737} & 0.834 & \textbf{0.050} & 0.077 & 0.122 & \textbf{0.151} \\
Eight (prototype III) & 0.75 & 0.143 & $4.69\pm0.71$ & 1.34 & 0.872 & 0.903 & 0.059 & \textbf{0.073} & \textbf{0.102} & 0.159 \\
\midrule
Gate (prototype II) & 0.65 & 0.212 & $8.20\pm0.49$ & 0.77 & 0.670 & 0.833 & 0.054 & 0.068 & 0.085 & 0.121 \\
\bottomrule
\end{tabular*}
\end{table*}

\begin{figure*}[t]
    \centering
    \includegraphics[width=0.98\textwidth]{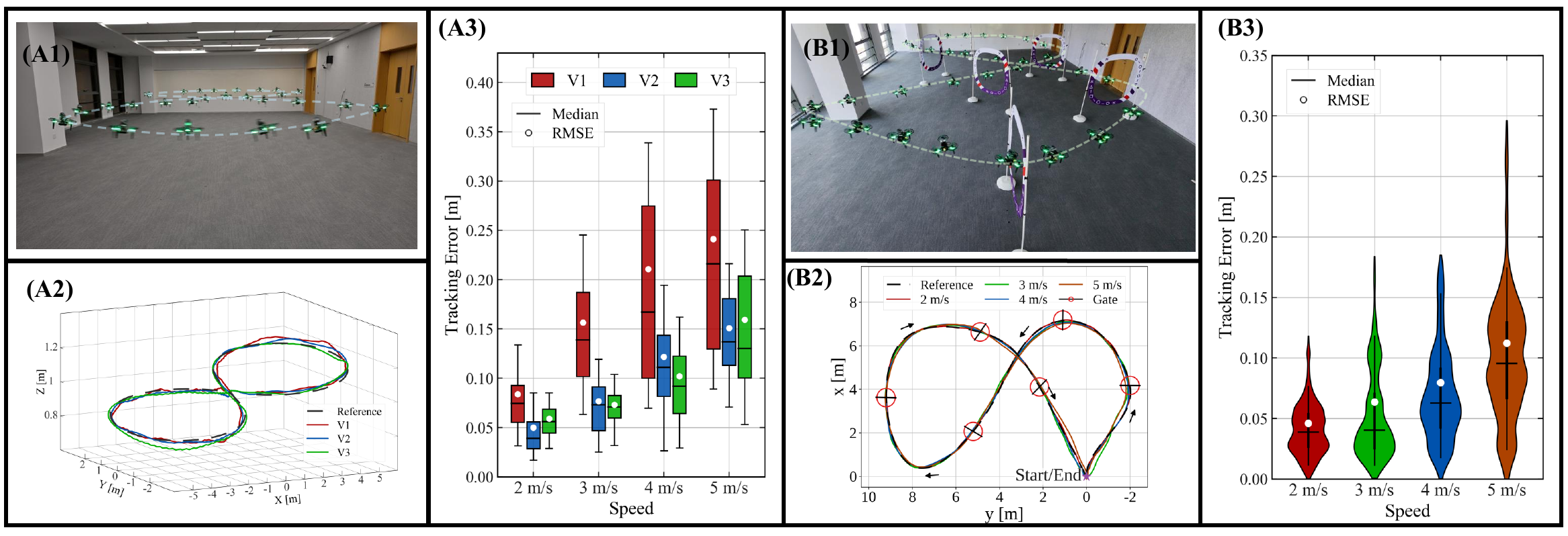}
    \caption{Trajectory tracking experiments for configuration selection and gate passing validation. A(1) shows overlaid snapshots during lemniscate flight. A(2) compares the reference and executed trajectories of the three prototypes on a three-dimensional lemniscate trajectory with a $10\,\mathrm{m}\times6\,\mathrm{m}$ footprint and a height range of $0.8\,\mathrm{m}$--$1.2\,\mathrm{m}$. A(3) reports distributions of tracking error at reference speeds from $2\,\mathrm{m/s}$ to $5\,\mathrm{m/s}$. B(1) shows the selected prototype II in gate passing, and B(2)--B(3) report its executed trajectories and distributions of tracking error.}
    \label{fig:tracking_exp}
\end{figure*}

\subsection{Configuration Selection and Agile Flight Performance}

We experimentally compare three prototypes with $\alpha=0.50$, $0.65$, and $0.75$, denoted as prototypes I--III, to evaluate the tradeoff between swept-FoV refresh and agile flight performance. Fig.~\ref{fig:tracking_exp}(A2)--(A3) show all prototypes following the lemniscate reference trajectory, but the tracking error distributions broaden as speed increases, especially for prototype I. This trend indicates that faster swept-FoV refresh increases compensation demand during agile translation. Prototype III maintains small errors at moderate speeds but has the longest rotation period, whereas prototype II keeps a relatively short period while remaining close to the best tracking accuracy across the speed range.

Table~\ref{tab:self_rotation_tracking} summarizes this tradeoff. As $\alpha$ increases, $\tau_{\mathrm n}$ decreases from $0.333$ to $0.143$, and the measured self-rotation rate during steady hover decreases from $11.03\,\mathrm{rad/s}$ to $4.69\,\mathrm{rad/s}$, confirming that $\alpha$ effectively sets the passive self-rotation operating point. The $\bar{\Omega}_z\pm\sigma$ values are computed from a selected $12\,\mathrm{s}$ steady hover window, and $T_s=2\pi/\bar{\Omega}_z$. The dynamic flight metrics include $e_{\mathrm{RMSE}}$ for position tracking. The $5\,\mathrm{m/s}$ metrics $\Delta f_{\mathrm{RMS}}$ and $U_{95}$ denote the RMS INDI correction to the NMPC thrust command and the 95th percentile of maximum rotor thrust utilization, respectively. At $5\,\mathrm{m/s}$, prototype I has the shortest period ($0.57\,\mathrm{s}$) but the largest tracking error ($0.241\,\mathrm{m}$) and INDI correction ($0.943\,\mathrm{N}$), whereas prototype III has the longest period ($1.34\,\mathrm{s}$) and the highest thrust utilization ($0.903$). Prototype II achieves $e_{\mathrm{RMSE}}=0.151\,\mathrm{m}$ with $\Delta f_{\mathrm{RMS}}=0.737\,\mathrm{N}$ and $U_{95}=0.834$, and is therefore selected for subsequent autonomous flight experiments.

To test prototype II in a gate-passing task, we further fly it through a sequence of gates at reference speeds from $2\,\mathrm{m/s}$ to $5\,\mathrm{m/s}$, as shown in Fig.~\ref{fig:tracking_exp}. The executed trajectories pass through the gates while following the reference, and the error distributions remain compact across speeds. At $5\,\mathrm{m/s}$, the RMSE is $0.121\,\mathrm{m}$ with $\Delta f_{\mathrm{RMS}}=0.670\,\mathrm{N}$ and $U_{95}=0.833$, confirming that the prototype II configuration preserves sufficient control margin in constrained agile flight.

\subsection{Self-Rotating Dynamics Model Validation}

The prototype-II coefficients used in the nominal prediction model are:
\begingroup
\setlength{\arraycolsep}{2.5pt}
\begin{gather*}
\bm{D}/m = \operatorname{diag}(0.307,0.325,0.145)\,\mathrm{s}^{-1},\\
\bm{c} = 10^{-3}[-23.38,12.06,35.94]^{\mathrm T}\,\mathrm{N\,m}.
\end{gather*}
\begin{gather*}
A_v = 10^{-3}
\left[
\begin{array}{rrr}
\phantom{1}-5.15 & -13.52 & -0.77\\
\phantom{-}14.77 & \phantom{-1}0.19 & \phantom{-}1.01\\
-14.08 & \phantom{1}-3.14 & \phantom{-}1.85
\end{array}
\right]\,\mathrm{N\,s},
\\[-0.2em]
B_{\omega} = 10^{-3}
\left[
\begin{array}{rrr}
\phantom{1}-0.77 & \phantom{1}-7.33 & \phantom{-}3.22\\
\phantom{1}-4.69 & \phantom{-1}1.27 & -1.01\\
\phantom{-1}3.63 & \phantom{-1}3.02 & -1.31
\end{array}
\right]\,\mathrm{N\,m\,s}.
\end{gather*}
\endgroup

\begin{table}[!t]
\caption{Held-Out Prediction Validation for Prototype II}
\label{tab:model_validation}
\centering
\begingroup
\small
\renewcommand{\arraystretch}{1.08}
\setlength{\tabcolsep}{3pt}
\begin{tabular*}{\columnwidth}{@{\extracolsep{\fill}}lcccc}
\toprule
\multicolumn{5}{c}{Translational acceleration RMSE $\bm{a}_B$ $\downarrow$ [$\mathrm{m/s^2}$]}\\
\midrule
Model & $x$ & $y$ & $z$ & Vec.\\
\midrule
Rigid body & 1.81 & 1.98 & 1.09 & 2.89\\
+ translational drag & 1.68 & 1.82 & 1.09 & \textbf{2.71}\\
+ drag + rotational residual & 1.68 & 1.82 & 1.09 & \textbf{2.71}\\
\bottomrule
\end{tabular*}
\vspace{0.3em}

\begin{tabular*}{\columnwidth}{@{\extracolsep{\fill}}lcccc}
\toprule
\multicolumn{5}{c}{Angular-acceleration RMSE $\dot{\bm{\omega}}$ $\downarrow$ [$\mathrm{rad/s^2}$]}\\
\midrule
Model & $x$ & $y$ & $z$ & Vec.\\
\midrule
Rigid body & 10.01 & 9.37 & 7.76 & 15.76\\
+ translational drag & 10.01 & 9.37 & 7.76 & 15.76\\
+ drag + rotational residual & 7.89 & 7.86 & 2.00 & \textbf{11.31}\\
\bottomrule
\end{tabular*}
\endgroup
\vspace{-2em}
\end{table}

Using the prototype-II coefficients, we evaluate NMPC prediction accuracy on lemniscate logs excluded from parameter identification, with reference speeds from $2\,\mathrm{m/s}$ to $5\,\mathrm{m/s}$. To avoid attributing input mismatch to model error, the logged thrust--moment corresponding to the rotor-thrust command after INDI compensation is used as the model input. Table~\ref{tab:model_validation} compares a rigid-body baseline, the same model with translational drag, and the full model. The acceleration labels are computed from logged state derivatives, where $\bm{a}_B$ denotes the velocity derivative transformed into the body frame and $\dot{\bm{\omega}}$ denotes the filtered angular-rate derivative recorded by the controller. Vector RMSE is computed over all pooled validation samples as the root mean square of the Euclidean norm of the prediction error over the three axes. The translational-drag term reduces the vector acceleration RMSE from $2.89\,\mathrm{m/s^2}$ to $2.71\,\mathrm{m/s^2}$. The rotational residual model reduces the vector angular-acceleration RMSE from $15.76\,\mathrm{rad/s^2}$ to $11.31\,\mathrm{rad/s^2}$, mainly by decreasing the yaw-axis RMSE from $7.76\,\mathrm{rad/s^2}$ to $2.00\,\mathrm{rad/s^2}$. This yaw-axis reduction is consistent with the yaw-dominant rotational mismatch associated with passive self-rotation. The identified drag and residual terms improve nominal NMPC prediction, leaving the remaining mismatch to INDI.

\begin{figure}[t]
    \centering
    \vspace*{8pt}
    \includegraphics[width=0.815\columnwidth]{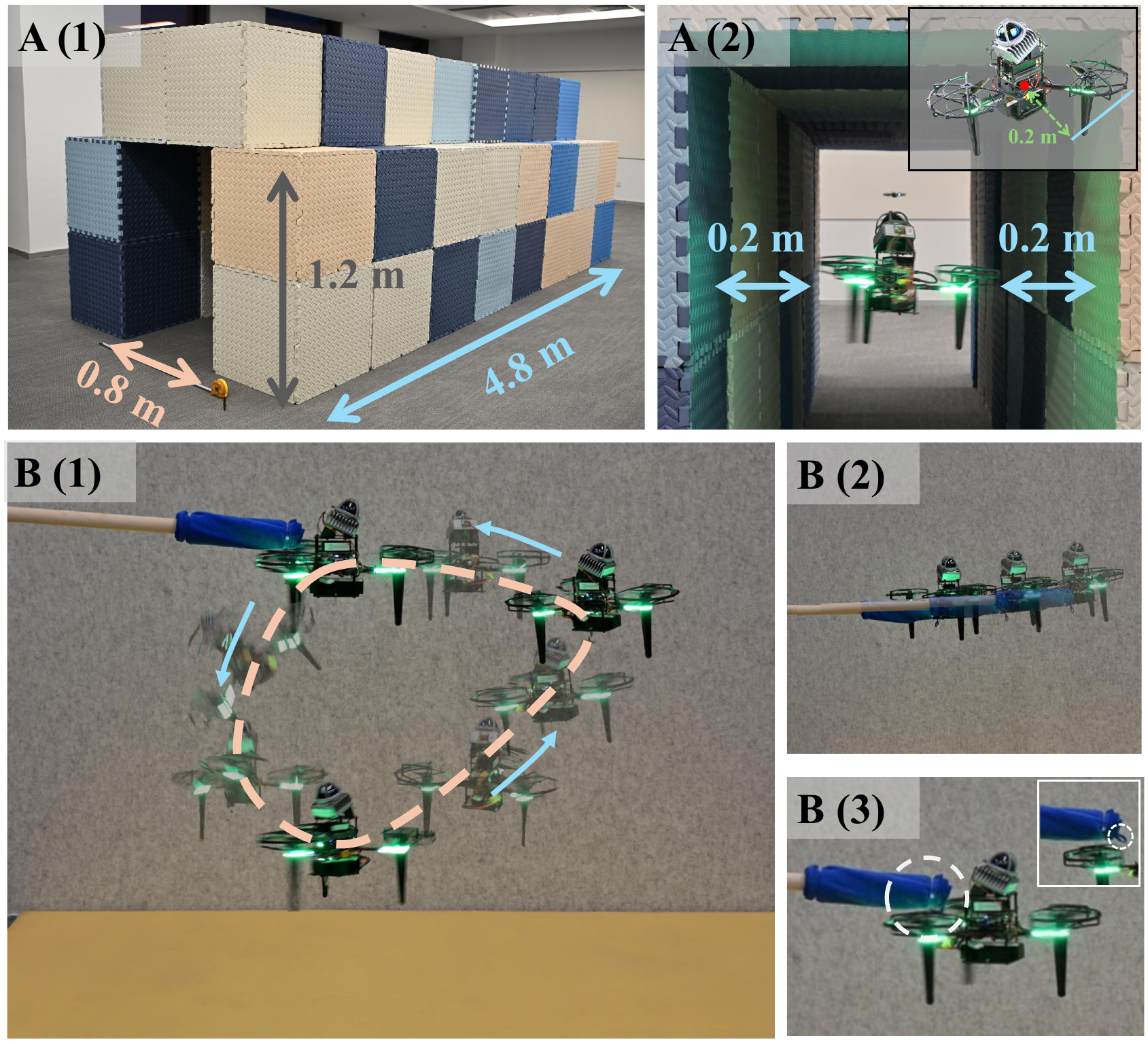}
    \caption{Robustness experiments. A(1) shows the confined passage with a height of $1.2\,\mathrm{m}$, width of $0.8\,\mathrm{m}$, and length of $4.8\,\mathrm{m}$. A(2) shows the tricopter flying through this confined passage at a reference speed of $3\,\mathrm{m/s}$ with a maximum reference acceleration of $4.5\,\mathrm{m/s^2}$. B(1)--B(3) show disturbance-recovery tests during hovering under cloth-strip contact with a rotating propeller and sustained airframe push.}
    \label{fig:disturbance_exp}
    \vspace{-0.5em}
\end{figure}

\subsection{Robust Flight under External Disturbances}

To evaluate closed-loop robustness beyond nominal tracking, prototype II is tested in confined-space flight and external-contact recovery. For confined-space flight, the vehicle flies through a foam tunnel of $0.8\,\mathrm{m}\times1.2\,\mathrm{m}\times4.8\,\mathrm{m}$ at a reference speed of $3\,\mathrm{m/s}$ with a maximum reference acceleration of $4.5\,\mathrm{m/s^2}$, as shown in Fig.~\ref{fig:disturbance_exp}(A). The vehicle has an approximate lateral half-span of $0.2\,\mathrm{m}$ from its center to the outermost side. In the $0.8\,\mathrm{m}$-wide tunnel, the estimated trajectory has a mean absolute deviation of $0.024\,\mathrm{m}$ from the passage centerline and a maximum deviation of $0.072\,\mathrm{m}$. In the disturbance-recovery test shown in Fig.~\ref{fig:disturbance_exp}(B), we apply two types of external contact to the hovering vehicle: brief contact between a cloth strip mounted on a wooden stick and a rotating propeller blade, and a sustained push on the airframe. In the brief propeller-contact trial, the strongest impact produces a maximum displacement of $0.62\,\mathrm{m}$ from the pre-contact hover position, including a vertical drop of $0.61\,\mathrm{m}$, and the vehicle returns to within $0.15\,\mathrm{m}$ in about $5\,\mathrm{s}$. In the sustained-push trial, the vehicle is displaced laterally by about $0.44\,\mathrm{m}$ and returns to within $0.15\,\mathrm{m}$ of the pre-push hover position in about $1.8\,\mathrm{s}$ after the displacement peak. In both contact tests, the vehicle remains controllable and recovers to stable hovering.

\begin{figure}[t]
    \centering
    \includegraphics[width=0.85\columnwidth]{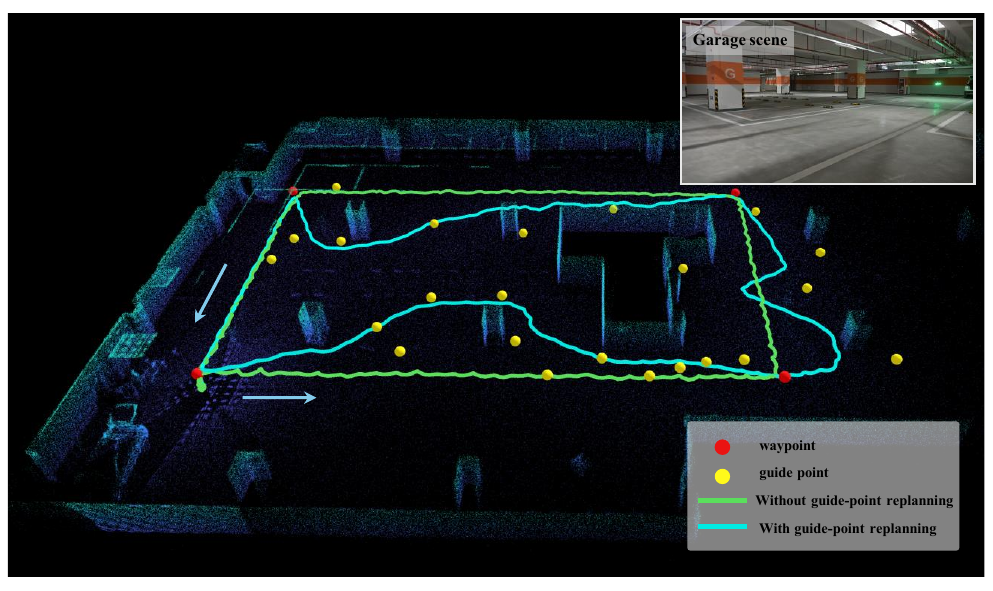}
    \caption{Underground-garage inspection experiment for guide-point evaluation. The inset shows a camera view of the experimental environment. The reconstructed point cloud is overlaid with sparse waypoints, selected guide points, and the executed trajectories with and without guide-point replanning.}
    \label{fig:garage_exp}
    \vspace{-1.5em}
\end{figure}

\subsection{Autonomous Navigation and Guide-Point Evaluation}

We first evaluate closed-loop navigation under passive self-rotation without guide-point selection. The test is performed in a $25\,\mathrm{m}\times25\,\mathrm{m}$ outdoor area with trees and hedges, as shown in Fig.~\ref{fig:forest_navigation}. Sparse waypoints provide only coarse route guidance, and five runs are conducted with reference speeds of $1$, $2$, $3$, $4$, and $5\,\mathrm{m/s}$, respectively. Prototype II replans locally around the trees and hedges and completes all five runs without collision under passive self-rotation.

To evaluate the effectiveness of guide-point replanning in waypoint-based inspection, autonomous navigation is performed with and without guide-point replanning in an underground-garage environment. As shown in Fig.~\ref{fig:garage_exp}, both trials use the same waypoint sequence in the same region. We quantify relative task-level coverage by voxelizing the two run point clouds at $0.1\,\mathrm{m}$ resolution and using their union as the reference set $\mathcal{V}_{\mathrm{ref}}=\mathcal{V}_{\mathrm{guide}}\cup\mathcal{V}_{\mathrm{no}}$. We report relative three-dimensional coverage $C_{\mathrm{3D}}=|\mathcal{V}_{\mathrm{run}}|/|\mathcal{V}_{\mathrm{ref}}|\times100\%$ and $3\,\mathrm{m}$ trajectory-neighborhood coverage $C_{3m}=|\{c\in\mathcal{C}_{\mathrm{ref}}\mid d(c)\leq3\,\mathrm{m}\}|/|\mathcal{C}_{\mathrm{ref}}|\times100\%$, where $\mathcal{C}_{\mathrm{ref}}$ is the projected reference map and $d(c)$ is the nearest trajectory distance. These metrics compare the two trials against this shared relative reference, rather than measuring absolute coverage of the complete garage environment.

As shown in Table~\ref{tab:garage_metrics}, the guide-point trial takes about $21.5\%$ longer, while increasing relative three-dimensional coverage from $51.9\%$ to $61.4\%$ and $3\,\mathrm{m}$ trajectory-neighborhood coverage from $55.9\%$ to $71.1\%$. To account for the longer execution time, we also compare time-normalized relative coverage rates. The guide-point trial achieves a comparable relative three-dimensional coverage rate ($0.57\%/\mathrm{s}$ versus $0.58\%/\mathrm{s}$), while slightly increasing the $3\,\mathrm{m}$ trajectory-neighborhood coverage rate from $0.63\%/\mathrm{s}$ to $0.65\%/\mathrm{s}$. For this route, guide-point replanning increases measured relative coverage while maintaining comparable time-normalized coverage rates despite the longer execution time.

\begin{table}[!h]
\caption{Underground-Garage Inspection Metrics}
\label{tab:garage_metrics}
\centering
\small
\renewcommand{\arraystretch}{1.25}
\begin{tabular*}{\columnwidth}{@{\extracolsep{\fill}} l c c c}
\toprule
Method & Time ($\mathrm{s}$) & $C_{\mathrm{3D}}$ $\uparrow$ & $C_{3m}$ $\uparrow$ \\
\midrule
Guide point & 108.7 & \textbf{61.4\%} & \textbf{71.1\%}\\
Direct waypoint & 89.4 & 51.9\% & 55.9\%\\
\bottomrule
\end{tabular*}
\vspace{-0.50em}
\end{table}

\section{Platform-Level Comparison}
\label{sec:platform_comparison}

\begin{figure}[t]
    \centering
    \vspace*{6pt}
    \includegraphics[width=0.826\columnwidth]{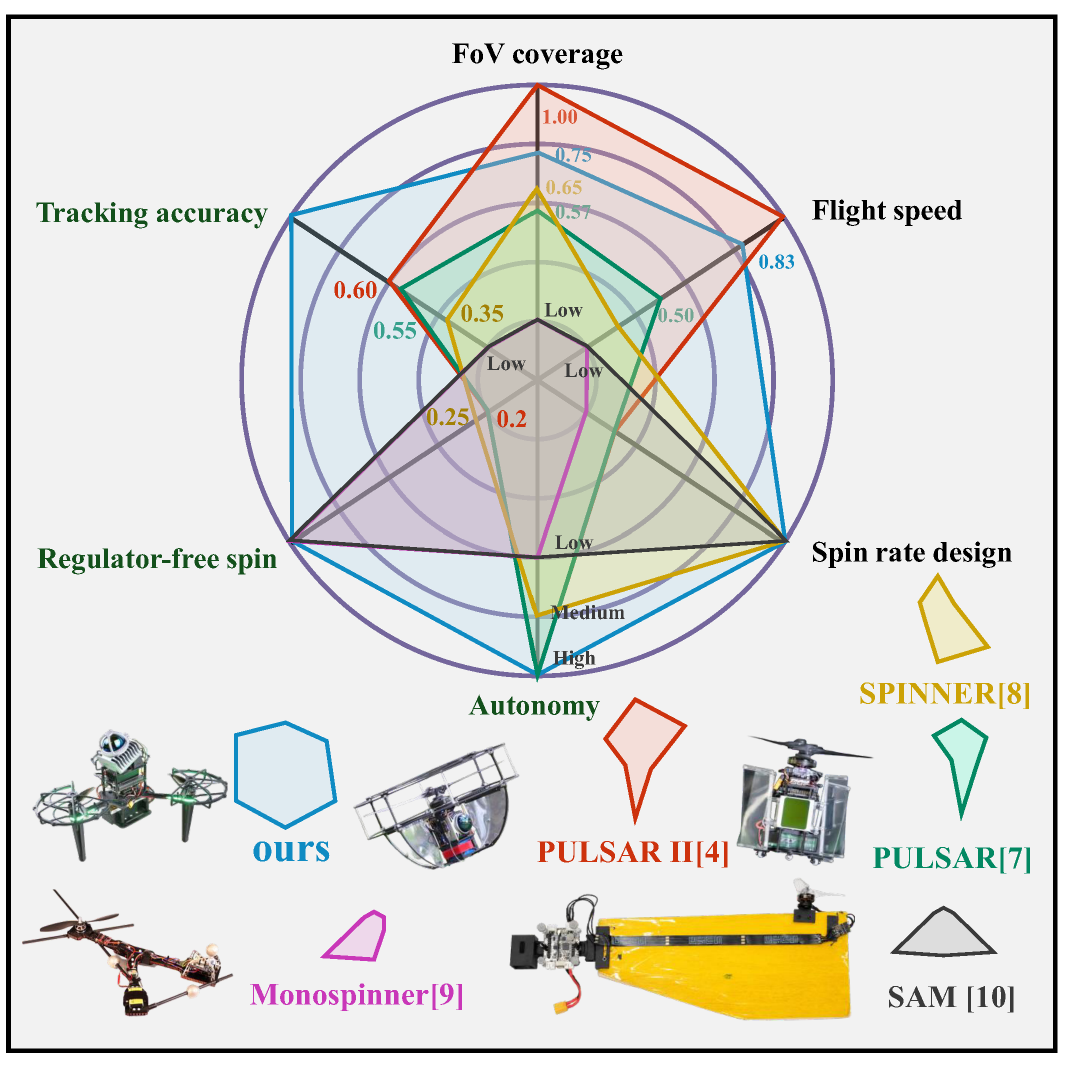}
    \caption{Platform-level comparison of representative self-rotating UAVs based on reported quantitative metrics and qualitative capability levels.}
    \label{fig:platform_comparison}
    \vspace{-1.5em}
\end{figure}

To clarify the architectural tradeoff behind the proposed tricopter, we compare it with representative self-rotating UAVs, including PULSAR II~\cite{chen2026pulsar2}, PULSAR~\cite{cai2023pulsar}, SPINNER~\cite{zhou2026spinner}, Monospinner~\cite{mueller2016monospinner}, and SAM~\cite{smooth2025monocopter}. Fig.~\ref{fig:platform_comparison} provides a normalized comparison based on reported metrics. FoV coverage is scored as $S_{\mathrm{FoV}}=\Omega_{\mathrm{eff}}/(4\pi)$, flight speed as $S_v=v_{\max}/\max_i v_{\max,i}$, and tracking accuracy at the speed closest to $2\,\mathrm{m/s}$ as $S_{\mathrm{track}}=e_{\min}/e_i$, where $e_i$ is the reported position-tracking error or RMSE of platform $i$, and $e_{\min}$ is the smallest comparable error. Autonomy is scored as High for onboard planning or replanning in cluttered or unknown environments, Medium for autonomous flight or tracking without autonomous planning, and Low when the capability is not reported or not directly comparable. Spin-rate design reflects the reported number of self-rotation operating points. Actuation simplicity is scored as $S_{\mathrm{act}}=1/N_a$, and regulator-free spin is scored as $S_{\mathrm{reg}}=1/(1+N_r)$, where $N_a$ is the number of propulsion actuators and $N_r$ is the number of additional anti-torque or spin-regulation surfaces.

The comparison indicates that the proposed tricopter achieves a strong overall profile among self-rotating UAVs, combining high tracking accuracy, high flight speed, onboard planning-based autonomy, and regulator-free passive self-rotation.

\section{Conclusion}

This letter presents a configuration-induced passively self-rotating tricopter for perception-enhanced autonomous flight. By treating the rear-arm configuration as an airframe-level design variable, the proposed platform regulates the passive self-rotation operating point to balance swept-FoV refresh rate and flight performance. The system combines a tilted LiDAR, a self-rotating flight model, an NMPC-INDI control framework, local replanning, and guide-point replanning for waypoint-based inspection. Real-world experiments show that the platform expands the vertical sensing envelope, tracks fast lemniscate trajectories, navigates cluttered environments, and, in the underground-garage inspection, records higher measured relative coverage in the guide-point trial. These results validate the proposed airframe-control-planning design and indicate that passive self-rotation can support autonomous navigation without additional sensor-actuation mechanisms.


\end{document}